\def\year{2020}\relax
\documentclass[letterpaper]{article} 
\usepackage{aaai20}  
\usepackage{times}  
\usepackage{helvet} 
\usepackage{courier}  
\usepackage[hyphens]{url}  
\usepackage{graphicx} 
\usepackage{graphicx}  
\usepackage{amsmath}
\usepackage{amssymb}
\usepackage[ruled,linesnumbered]{algorithm2e}
\usepackage{subfigure}
\usepackage{mathrsfs}
\usepackage{multirow}
\usepackage{multicol}
\usepackage{xcolor}

\newcommand{\js}[1]{\textcolor{black}{#1}}
\newcommand{\lyx}[1]{\textcolor{black}{#1}}
\usepackage[american]{babel}

\newcommand{\mca}[1]{\mathcal{#1}}
\newcommand{\mbf}[1]{\mathbf{#1}}
\newcommand{\floor}[1]{\lfloor #1\rfloor}
\newcommand{\ceil}[1]{\lceil #1\rceil}

\title{Finding Action Tubes with a Sparse-to-Dense Framework}
\author{
\Large \textbf{Yuxi Li\textsuperscript{\rm 1, \rm 2}, Weiyao Lin\textsuperscript{\rm 1, \rm 2}\thanks{∗Corresponding Author, Email: wylin@sjtu.edu.cn}, Tao Wang\textsuperscript{\rm 1}, John See\textsuperscript{\rm 3}} \\ \Large \textbf{Rui Qian\textsuperscript{\rm 1}, Ning Xu\textsuperscript{\rm 4}, Limin Wang\textsuperscript{\rm 5}, Shugong Xu\textsuperscript{\rm 2}} \\ 
\textsuperscript{\rm 1}School of Electronic Information and Electrical Engineering, Shanghai Jiao Tong University, China\\ 
\textsuperscript{\rm 2}Shanghai Institute for Advanced Communication and Data Science, Shanghai University, China\\
\textsuperscript{\rm 3}Multimedia University, Malaysia\\
\textsuperscript{\rm 4}Adobe Research, USA\\
\textsuperscript{\rm 5}State Key Laboratory for Novel Software Technology, Nanjing University, China\\
}
 
\begin{document}

\maketitle

\begin{abstract}
The task of spatial-temporal action detection has attracted \js{increasing} attention among researchers. \js{Existing} dominant methods solve this problem by relying on short-term information and \js{dense serial-wise} detection on each \js{individual} frames or clips. 
   \js{Despite their effectiveness,} these methods showed \js{inadequate use}  
   of long-term information and \js{are prone to} inefficiency. 
   In this paper, \js{we propose for the first time,} 
   \js{an efficient framework that generates action tube proposals from video streams} \lyx{with a single forward pass} 
   in a sparse-to-dense manner. There are \js{two key} characteristics in this framework: (1) Both long-term and short-term sampled information are explicitly utilized in our spatio-temporal network, (2) A \js{new} dynamic feature sampling module (DTS) is designed to effectively approximate the tube output while keeping the system \js{tractable}. We 
   \js{evaluate the efficacy} of our model on the UCF101-24, JHMDB-21 and UCFSports \js{benchmark} datasets, \js{achieving} 
   promising results \js{that are} competitive to state-of-the-art methods.
   \js{The proposed} \lyx{sparse-to-dense} \js{strategy} \js{rendered} our framework \js{about $7.6$ times more efficient than the nearest competitor.} 
\end{abstract}

\begin{figure}[t]
  \centering
  \subfigure[Pipeline of previous works]{
  \includegraphics[width=0.4\textwidth]{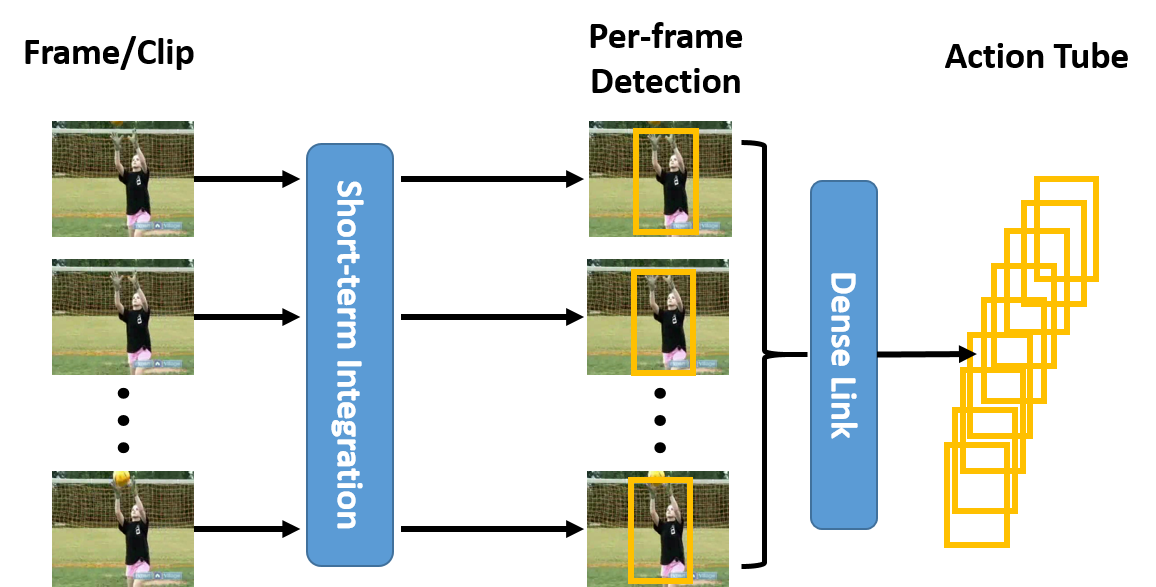}\label{fig:concept-previous}}
  \vspace{-1mm}
  \subfigure[Pipeline of our work]{
  \includegraphics[width=0.4\textwidth]{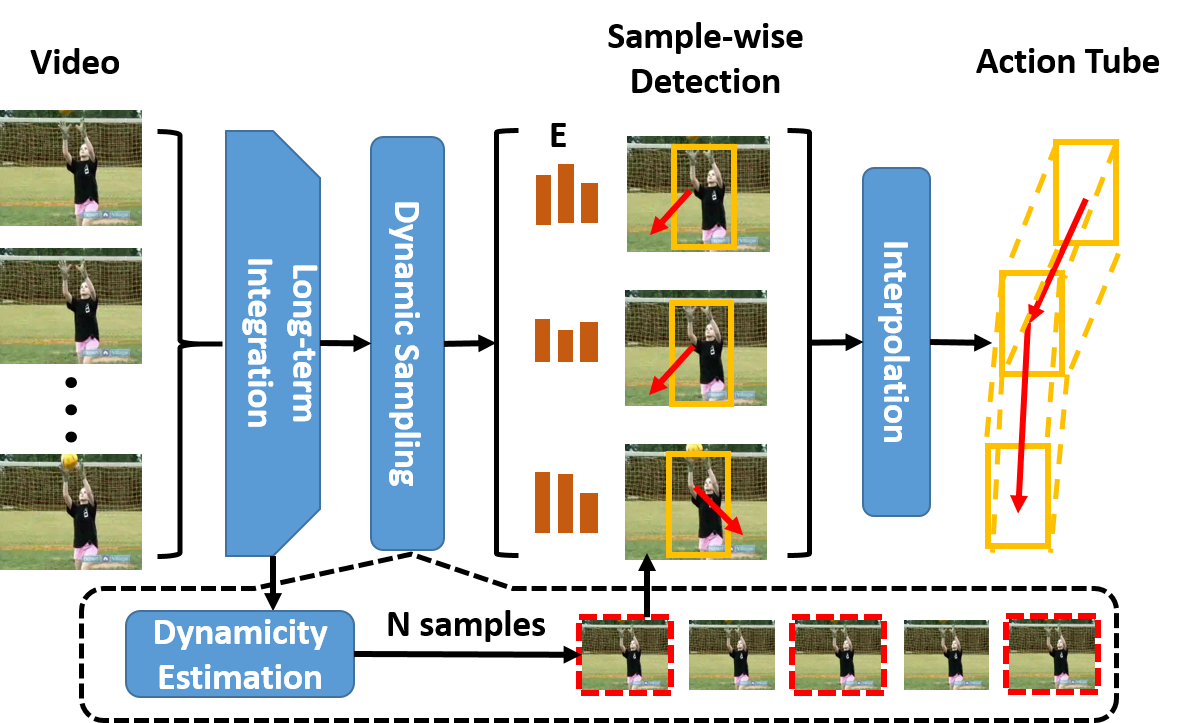}\label{fig:concept-ours}}
  \caption{Comparison between previous pipelines and our pipeline}
  \label{fig:concept}
\end{figure}

\section{Introduction}
Spatial-temporal action detection is an essential technology for video understanding applications. In contrast to action recognition or temporal localization, where a video-level class label or temporal proposals are to be assigned 
spatial-temporal localization involves three sub-tasks: determining the interval of action occurrence,
localizing the actors within the interval and correctly categorizing the action. This makes for an extremely complex task. 

 \lyx{To the best of our knowledge, previous} approaches~\cite{peng2016multi,saha2016deep,singh2017online,hou2017end,yang2017spatio} \lyx{conform to a dense detection paradigm}
 (Fig. \ref{fig:concept-previous}), which detects dense bounding boxes with actor detectors~\cite{liu2016ssd,Ren_2017} first on \lyx{with short-term information from} frame or snippet level and then link the proposals together via certain heuristic linking algorithms. Relying on powerful convolution neural networks, these methods achieved good performance in both localization and classification. However, there are several shortcomings
 of such pipelines. Firstly, the input of most methods only contain brief temporal information, which makes it hard to identify actions that are somewhat similar through most parts of the
 video, e.g. both the action ``Long Jump" and ``Pole Vault" involve actors running at the beginning and are only different at the last part of these videos. Secondly, by simple but brute force means, it is inefficient to generate dense frame level proposals considering the computational complexity. 
 
  
 
 With these consideration, 
 we propose an \lyx{novel} framework for spatial-temporal action detection in a sparse-to-dense manner. In contrast to the pipelines of previous works, we first generate box proposals at sparsely sampled frames (Fig.~\ref{fig:concept-ours}), then we get the dense tube by interpolating the sparse proposals across a given detected time interval. 
 
 We solve the issues of previous works in two aspects. Firstly, we introduce a long-term feature augmentation module (LFA) to combine both the long-term and short-term information in a single forward pass. We exploit this combined feature to generate more reliable spatial and temporal proposals. Additionally, the augmented feature is further utilized to generate embedding and shift vectors (symbolized by `E' and red arrows in Fig.~\ref{fig:concept-ours}) for each bounding box, facilitating the tube generation process. 

Secondly, we propose an adaptive dynamic temporal sampling module (DTS) to achieve sparse detection and decrease computation complexity. We argue that if the number of sampled frames is sufficient and the detection on each sampled frame is accurate, then the interpolation result over each bounding box forms an approximated tube proposal, as shown in Fig.~\ref{fig:concept-ours}. Additionally, 
we hypothesize 
that the number of sampled frames required to achieve good approximation varies and depends on the dynamic extent of actions, e.g. when actors stand still and do not move, then the bounding box stays the same over time, therefore boxes on two sampled frames are sufficient to reconstruct the action tube. 
Based on these observation, we guide our network to estimate the level of action dynamics, or \emph{dynamicity} of each tube proposal, and 
DTS adaptively maps 3D features to 2D form for further detection. With the DTS module, our framework can achieve a good trade-off between model complexity and accuracy of localization.

In summary, our contributions are three folds. 
\begin{enumerate}
    \item We propose an \lyx{general} framework for the task of spatial-temporal action detection. In this framework, tube proposals are generated by a \emph{sparse-to-dense} mechanism with a single forward pass. Competitive results on three benchmark datasets are obtained with an inference speed that is $7.6$x faster than the nearest competitor.
    
    \item \lyx{In this framework}, we design a long-term feature augmentation module (LFA) for enhanced feature representation. We further exploit the augmented feature to generate reliable proposals and auxiliary vectors for box association.
    
    \item \lyx{For sparse detection}, we introduce an adaptive sampling module called dynamic temporal sampling (DTS) to boost the accuracy of tube construction at a much lower computation cost.
\end{enumerate}

\section{Related works}


\textbf{Action recognition.} Owing to its successful application of deep neural networks for image classification~\cite{simonyan2014very,simonyan2014very,huang2017densely}, some works have started to focus on the utilization of CNNs for video action recognition. Simonyan \& Zisserman~\cite{simonyan2014two} proposed a two-stream framework, which used two networks to extract features from appearance and optical flow input respectively, and combined them to obtain the final decision. ~\cite{Tran_2015} developed a 3D ConvNet to automatically extract the feature representation for actions. The I3D network \cite{carreira2017quo} further inflated the networks pre-trained on ImageNet (2D) \cite{deng2009imagenet} to form an efficient 3D network for action recognition. Despite these works achieving good results on standard benchmarks, these methods only focused on labels at video level and is not capable of providing more detailed information as when the action starts or how the actors move.

\textbf{Temporal action detection.} Some early works on action detection mainly focused on the temporal localization of video actions, which aims at finding the time interval of action instances. A few recent works~\cite{lin2017single,zhao2017temporal,chao2018rethinking} extracted frame level deep features and apply either RPN or pyramidal structure for 1-dimensional action detection. In \cite{xu2017r,zhang2018s3d}, a 3D convNet was first utilized to extract 3D features for the proposal network or action detectors~\cite{Ren_2017,liu2016ssd}. However, none of these approaches yielded precise spatial-temporal tubes for action instances.

\textbf{Spatial-temporal action detection.} The earliest pipeline devised for action tube detection was proposed in \cite{Gkioxari_2015}, where the R-CNN structure~\cite{girshick2014rich} was applied on each frame for action bounding box detection and the results were linked by viterbi algorithm; however, it cannot determine the interval of actions. \cite{saha2016deep} addressed the problem by introducing an extra labeling operation after linking. The following methods mainly strove for better feature representation by involving contextual information \cite{peng2016multi}, extracting spatial-temporal features from short time snippets \cite{saha2017amtnet,kalogeiton2017action,hou2017end}, or resorting to recurrent neural networks~\cite{huang2018online}. All of these works follow a common detect-and-link framework, the output spatial proposals are linked either by viterbi algorithm or other improved strategies~\cite{singh2017online,huang2018online}. Such  pipelines require \lyx{dense detections} for each video. This inefficiency worsens when optical flow computation is taken into account. Further, the input model only contains local short time information and can be ambiguous during some short intervals.

In contrast to the works mentioned above, our framework is the first \lyx{pipeline to generate action tube proposals in a sparse-to-dense manner and handles both long and short term information in a single forward pass }. Long-term temporal context is considered by capturing characteristics of action instances for final association. Besides, the dynamic temporal sampling (DTS) module further decreases the search space to achieve good approximation of the ground truth bounding boxes. 

\begin{figure*}[htb]
\centering
\includegraphics[width=0.9\textwidth]{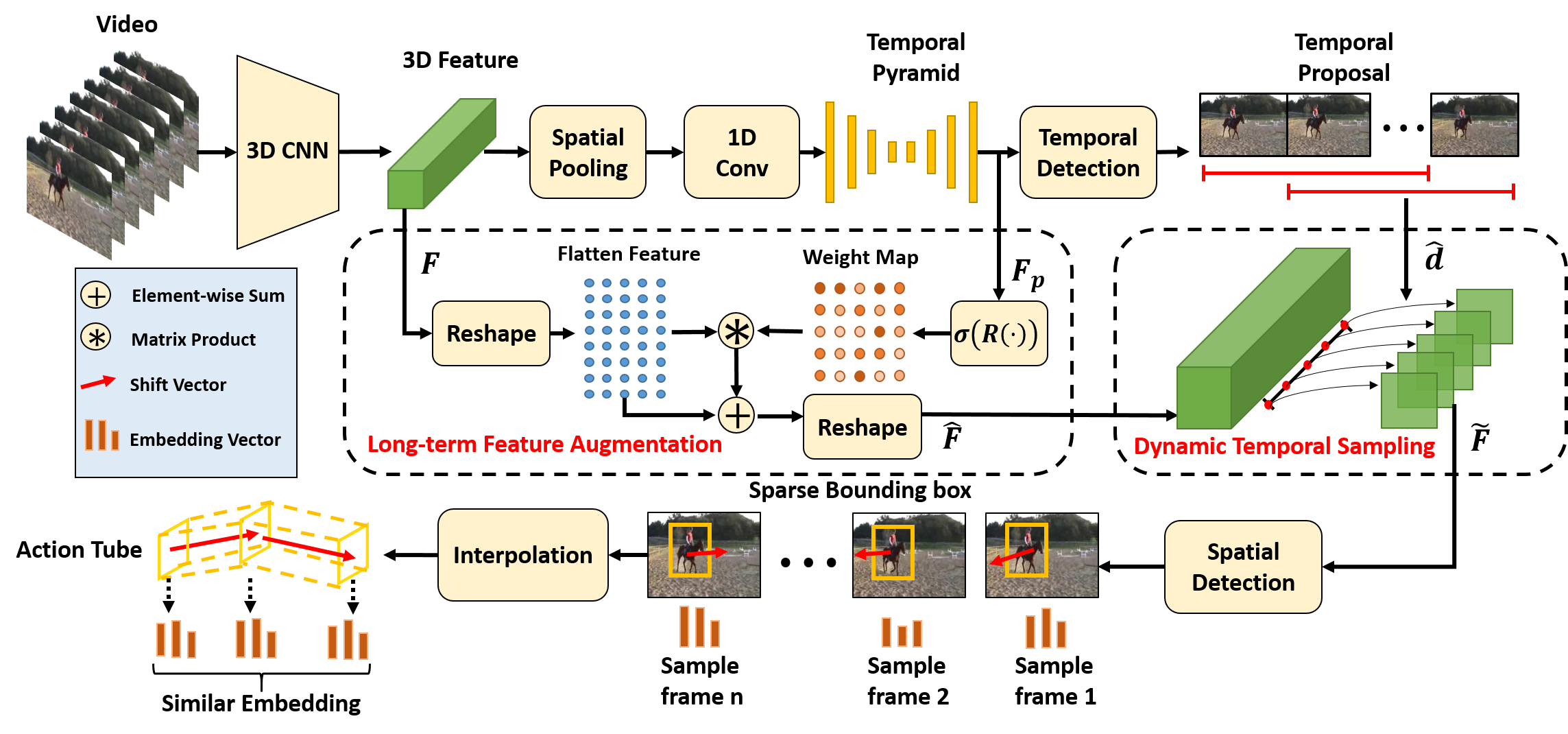}
\caption{Overview of the proposed framework.}\label{fig:overview}
\end{figure*}
\section{Method}
 Given a video $\mathcal{V}$, our goal is to find a tube set $\mca{A}=\{\Phi_j|j=1\dots N\}$, where the symbol $\Phi_j$ denotes a certain action tube. For each tube $\Phi_j$, a class label $c_j$ and a spatial bounding box set $\mca{B}_j=\{b_t|t\in [s_j, e_j]\}$ is assigned to it, where $b_t$ denotes the four dimensions of the bounding box and $s_j, e_j$ denote the starting and ending frame indices.

Our framework is illustrated in Fig.~\ref{fig:overview}. At first, an input video stream is temporally sampled to form a fixed-length tensor of $T$ frames. The tensor is fed into a 3D ConvNet for feature extraction. The network transforms the feature into a 1D temporal feature pyramid to capture long-term information and generate temporal proposals. \lyx{Since the original 3D features have limited temporal receiptive field}, the following long-term feature augmentation module (LFA) boosts the 3D features with long-term weighted recombination. Next, the dynamic temporal sampling module (DTS) takes the augmented feature and temporal proposals to generate sparse 2D feature samples over time. For each temporal proposal,the sampled features are utilized to generate spatial bounding boxes, which are accompanied with embedding and shift vectors. With the guidance of these vectors, the boxes are \lyx{associated and interpolated to form final tube proposals along corresponding temporal proposal.}

\subsection{Temporal long-term integration}
To extract long-term information and integrate it with short-term features, we design a temporal pyramid and long-term augmentation module to handle features at temporal scale.

\textbf{Temporal pyramid and proposals.} The construction of the temporal pyramid structure is illustrated in Fig.~\ref{fig:overview}. We first squeeze the input 3D feature into a single dimension via spatial average pooling. To obtain 1D temporal features, we first shrink the temporal resolution by convolution with stride $2$, and then reversely enlarge the temporal size with deconvolution operators. These manipulations construct a two-path temporal feature pyramid, i.e. downsample path and upsample path.

We then predict temporal proposals from the pyramid. Similar to \cite{zhang2018s3d}, on each level of the upsample path, we assign the pixels with a set of temporal anchors of different scales. For each anchor, the temporal detector predicts the \lyx{actioness} score and regresses the offsets relative to the center and length of it via 1D temporal convolution. 


\textbf{Long-term feature augmentation.} The LFA module works by explicitly recombining long-term information via an attention mechanism and fuses it with short-term features extracted earlier. The details are illustrated in Fig.~\ref{fig:overview}. 
We take the feature $\mca{F}_p$ from the last layer of upsample path (of size $C\times T_f$) as the output of temporal pyramid, where $C$ is the channel number and $T_f$ is the temporal size of feature. The model learns a mapping function $\mca{R(\cdot)}$ that maps $\mca{F}_p$ into a subspace and normalizes it with a sigmoid function $\sigma(\cdot)$,
\begin{equation}
    \mca{S} = \sigma \left(\mca{R}(\mca{F}_p;\theta)\right)
\end{equation}
where $\mca{S}$ is a weight map of size $T_f \times T_f$, and $\theta$ represent the learnable parameters. In implementation, the function $\mca{R(\cdot)}$ is constructed by stacking three 1-D convolutional blocks followed by batch normalization and ReLU function. Finally, we obtain the new 3D feature representation via weighted combination and fusion process in terms of Eq.~\ref{eq:attention}, which can be simply accomplished via matrix multiplication and a residual connection:
\begin{equation}\label{eq:attention}
    \hat{\mca{F}}_{c,t,h,w} = \mca{F}_{c,t,h,w} + \frac{1}{T_f}\sum_{k=1}^{T_f}{\mca{F}_{c,k,h,w}\mca{S}_{t,k}}
\end{equation}
where the tensor $\mca{F}$ is the original 3D feature input and $T_f$ is its temporal size, $t, h, w$ are the temporal and spatial coordinates of each pixel in feature map. Note that Eq. \ref{eq:attention} can be regarded as a recombination of each temporal component of input feature. At each time index of the output feature, the component is a weighted combination of all other time indices. With this module, we can guarantee that the time component of each 3D feature is encoded with long-term information.

We argue that our design of the long-short term integration mechanism has the following advantages: (\romannumeral1) Our network localizes the temporal interval at an early stage, thus avoiding the temporal labelling process~\cite{saha2016deep} used in other works as a post-processing step. 
(\romannumeral2) The feature pyramid guides the long-term recombination and fusion of original 3D features via LFA, which leads to more representative 3D features. The augmented features can be utilized to generate better detection results. 

\subsection{Dynamic temporal sampling}
The next step is to temporally sample the augmented 3D feature into 2D form for subsequent sparse detection. We design an adaptive dynamic temporal sampling (DTS) module to obtain sufficient feature samples at an appropriate sampling rate. 

The key mechanism of DTS is to generate an appropriate ground-truth dynamic level $d$ to guide the prediction of $\hat{d}$. Intuitively, for a scalar function $f(x)$, if its sampled values $Y=\{f(x_t)|t=1\dots\}$ are sufficient, then the piecewise interpolation between $(x_i, f(x_i))$ and $(x_{i+1}, f(x_{i+1}))$ is a good approximation to the corresponding original segment $f(x)$. Based on this intuition, we define $d$ as the minimum number of uniform samples such that the interpolation between samples form a tube as close as possible to the ground-truth. This process is depicted in Algorithm \ref{alg:dynamic}.
\begin{algorithm}[t!]
\caption{Ground-truth dynamic level generation}\label{alg:dynamic}
\small
\KwIn{Tube $\Phi$, threshold $\epsilon$}
\KwOut{dynamic level $d$}
get bounding box set $\mca{B}=\{b_i|i=1\dots T\}$ from tube $\Phi$ \;
\For{$d=1$ to $T$}
{$r=0$\;
uniformly sample $d$ bounding boxes from $\mca{B}$\ and obtain sampled box set $\mca{P}=\{b_{t_k}|{t_k}=\floor{kT/d}, k=1\dots d\}$\;
Interpolate over $\mca{P}$ to obtain $\hat{\mca{B}}=\{\hat{b}_i|i=1\dots T\}$\;
\For{$k=1$ to $T$}{$r=r+IOU(b_k,\hat{b}_k)$\;}
$r=r/T$\;
\If{$r\ge \epsilon$}{\Return{$d$\;}}
}
\end{algorithm}

\lyx{
To obtain an appropriate estimation, for each temporal proposal, we
estimate the dynamic level $\hat{d}$ via an additional convolution layer by considering the intensity of motion variation. During training stage, for each temporal proposal that matches a ground-truth tube, we assign the corresponding $d$ to it and train with a weighted smooth L1 loss \cite{Ren_2017}:
\begin{equation}
    L_d = smooth\_L_1(d-\hat{d};\gamma)
\end{equation}
where the negative part of loss function is suppressed by a factor of $\gamma$. Since having insufficient samples (small $n$ value) is more harmful to final approximation than large ones, we place a stronger penalty on the positive side of loss function. }

During inference time,  we set the sample number $n=\min(\ceil{\hat{d}}, N_{max})$, where the hyperparameter $N_{max}$ avoids sample numbers that are too large. Given a normalized temporal proposal $(s, e)$, where $s$ and $e$ are the starting and ending time indices, the 2D spatial features can be sampled from 3D features via linear interpolation:
\begin{equation}
\begin{split}
    \widetilde{\mca{F}}^i_{c,h,w} = & \sum_{k=1}^{T_f}{\hat{\mca{F}}_{c,k,h,w}}\max\left(0,1-\left|s+\frac{e-s}{n-1}i-k\right|\right)
\end{split}
\end{equation}
where $\widetilde{\mca{F}}^i$ is the $i$-th 2D feature map output.
Compared with dense frame-level detection, the design of DTS is able to decrease the number of detection operations to a large extent and make the framework more efficient during inference.

\subsection{Sparsely sampled bounding box detection}

\begin{figure}[htb]
    \centering
    \includegraphics[width=0.35\textwidth]{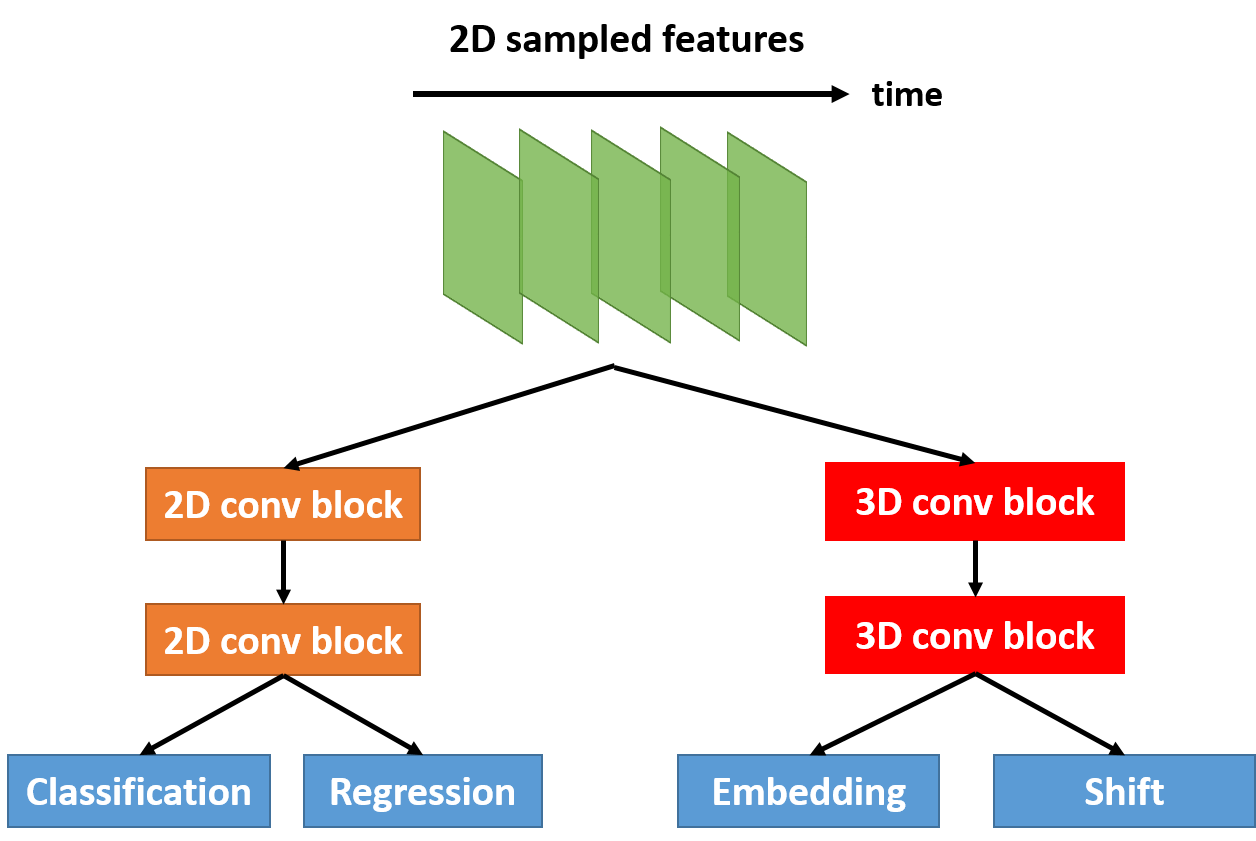}
    \vspace{3mm}
    \caption{Illustration of spatial detection network. Each convolutional block comprises of a convolutional layer, a batch normalization layer and a ReLU layer.}
    \label{fig:sd}
\end{figure}

Given the sparsely sampled 2D features, we detect actor bounding boxes on each feature map with a spatial detector network. The spatial detector network consists of two \lyx{branches}: the box detection \lyx{branch} and the association \lyx{branch} (shown in Fig. \ref{fig:sd}). In the box detection \lyx{branch}, we apply 2D convolution over each sampled feature map to obtain classification score and regress coordinate offsets, which is similar to \cite{liu2016ssd}. For the association branch, 3D convolution is applied on the stacked sampled features to capture the temporal relation among adjacent key frames, at the end of the path, the network outputs an embedding vector $\mbf{f}_i$ and a potential shift vector $\delta_i = [d_{x,i}, d_{y,i}]^T$ for each anchor box $b_i$. These vectors encode the appearance and spatial relationship between boxes in sparsely sampled frames of the same action instance.

Intuitively, the boxes belonging to the same action instance are expected to exhibit similar appearance feature, while the potential shift vector should indicate the direction and distance in which the bounding box of the current sample could shift to the corresponding box in the next sample. With this consideration, 
we design two losses $L_{a}$ and $L_{m}$ to regularize the \lyx{embedding and shift} vectors of positive anchors. To be specific, we sample the corresponding ground-truth tubes with the same sampling rate as the feature and treat the $i$-th sampled bounding box $b_{t_i}=(x_{t_i},y_{t_i},w_{t_i},h_{t_i})$ from tube $\Phi$ as the ground-truth for the $i$-th sampled feature map. The loss term $L_{a}$ guides the form of embedding vectors to optimize the cluster-like distance in feature space:
\begin{equation}
    L_{a} = \sum_{i}{C_{i}\mbf{d}_2(\overline{\mbf{f}},\mbf{f}_i})+(1-C_{i})[\alpha-\mbf{d}_2(\overline{\mbf{f}},\mbf{f}_i)]_+
\end{equation}
where $C_{i}$ is $1$ if anchor $i$ matches current ground-truth action instance, otherwise it equals $0$. $\mbf{d}_2(\cdot, \cdot)$ is the squared $L_2$ distance, the symbol $\overline{\mbf{f}}$ is the mean embedding vector over all anchors matching the instance.  $[\cdot]_+$ is the ReLU operator and $\alpha$ is a hyperparameter to control the margin between different cluster centers. 

The other loss term $L_{m}$ guides the prediction of box offsets to the next ground-truth sample, which is expressed as follows:
\begin{equation}
    L_{m} = \sum_{k=2}^{n}{\sum_{i}{C_{i}\mbf{d}_2(\delta_i, \delta_{gt,k})}}
\end{equation}
where $n$ is the number of sampled features, and $\delta_{gt,k}=[x_{t_{k+1}}-x_{t_k}, y_{t_{k+1}}-y_{t_k}]^T$ indicates the offset from the center of current sampled box to the next one.

\subsection{Dense tube generation from sparse proposals}
While the preceding steps predict the time interval of actions and their associated sparse bounding box proposals, the final step combines these proposals to form the dense action tube along the temporal dimension. In this scenario, we simply design a greedy strategy to generate tube proposals.

Given a temporal proposal $(s, e)$ and its confidence probability $p$, we can obtain $n$ sets of bounding boxes via DTS and the spatial detection network, which is denoted as $\mca{U}=\{\hat{\mca{B}}_t|t=1\dots n\}$, where $\hat{\mca{B}_t}$ is the box proposal set from the $t$-th sampled 2D feature. Note that in Fig. \ref{fig:sd}, the spatial detection network does not only predict the boxes, but also assign embedding vector and shift vectors for each anchor. We hereby define $p_{tj}$ as the probability score for class $c$ from box proposal $\hat{b_{tj}} \in \hat{\mca{B}_t}$, its corresponding embedding vector as $\mbf{f}_{tj}$ and potential shift vector as $\delta_{tj}$. For each pair of bounding boxes from adjacent samples, we compute their appearance distance $D_{a,tij}$ and spatial distance $D_{s,tij}$ at $t \in [1,n-1]$ between $\hat{b}_{ti} \in \hat{\mca{B}}_t$ and $\hat{b}_{(t+1)j} \in \hat{\mca{B}}_{t+1}$ as:
\begin{equation}
\begin{split}
D_{a,tij} &= ||\mbf{f}_{ti}-\mbf{f}_{(t+1)j}||_2 \\
D_{s,tij} &= ||\delta_{ti}-\overline{\delta}_{tij}||_2
\end{split}
\end{equation}
where $\overline{\delta}_{tij}$ denotes the center offsets from box $\hat{b}_{ti}$ to box $\hat{b}_{(t+1)j}$. We formulate the connectivity between the two adjacent boxes as:
\begin{equation}
    \mca{C}\left(\hat{b}_{ti},\hat{b}_{(t+1)j}\right) = \exp\left(-\frac{D_{a,tij}+D_{s,tij}}{2}\right)
\end{equation}
Selection of box proposals begin by first picking the starting box from set $\hat{\mca{B}}_1$ according to the classification probability:
\begin{equation}
     \hat{b_1}=\arg\max_{\hat{\mca{B}}_1}{p_{1j}}
\end{equation}
Subsequently, for the rest of the box proposal sets, we select boxes with the largest connectivity to the last box of current tube.
\begin{equation}
    \hat{b}_{(t+1)}=\arg\max_{ \hat{\mca{B}}_{(t+1)}}{\mca{C}\left(\hat{b}_{t},\hat{b}_{(t+1)j}\right)}
\end{equation}
Finally, we perform linear interpolation between contiguous boxes in sample set $\{\hat{b}_t\}$ to reconstruct the tube. The sampling rate for interpolation can be expressed as $f_s = \frac{n}{(e-s)T}f_v$,
where $f_v$ is the sample frequency from original raw video to fixed length tensor. After the association and interpolation steps, we obtain the dense frame level bounding boxes for each action tube. The score of the action tube is the average score across all predicted boxes multiplied by the \lyx{actioness} score $p$ of the temporal proposal.



\section{Experimental results}

\subsection{Experimental settings}

\textbf{Datasets.} We conduct our experiment on three common datasets -- UCF101-24, UCFSports and JHMDB-21 datasets. The UCF101-24 dataset~\cite{soomro2012ucf101} contains 3,207 untrimmed videos with frame level bounding box annotations for 24 sports classes. The dataset is challenging due to the frequent camera shake besides the actor movements. Following previous works~\cite{saha2016deep}, we report results for the first split. The JHMDB-21 is a subset of HMDB-51 dataset~\cite{Jhuang2013}, which contains a total of 928 videos with 21 types of actions. All video sequences are temporally trimmed. The results are reported as the average performance over 3 train-test splits. The UCFSports dataset~\cite{ucfsports} contains 150 videos of 10 sport action classes. We report the result on the standard split. Note that although the last two datasets are trimmed temporally, their samples are still suitable for our framework as they comprise of actions spanning the whole video.

\textbf{Metric.} We adopt the standard video-mAP (v-mAP)~\cite{Gkioxari_2015} as our metric for spatial-temporal action detection on all three datasets. A proposal is regarded as positive only when its tube overlap with an undetected ground-truth is larger than threshold $\Delta$. 

\textbf{Implementation details.}
Our model is implemented on an NVIDIA Titan 1080 GPU. We use the I3D network \cite{carreira2017quo} as our 3D feature extractor. 
During training, we use length $T$ of videos sampled to $96$ frames for UCF101, $32$ frames for JHMDB-21 and $48$ frames for UCFSports. We set the hyperparameters as: $\alpha=2, \epsilon=0.7, \gamma=0.1$. To avoid too large sampling points in DTS, we set the maximum sampling points as $10$ for UCF101-24, $4$ for JHMDB and $8$ for UCFSports. We train the network in two stages: First, we fix the weights in the spatial detection network and LFA, then train the backbone  
and temporal detection network 
for the purpose of learning the proposals and dynamic levels.
Next, we jointly train the network end-to-end 
to learn the final action tubes.
We use the SGD solver and train our network with an accumulative batch size of $16$.


    
    

\begin{figure*}[htb]
\centering
\includegraphics[height=0.35\textheight,width=\textwidth]{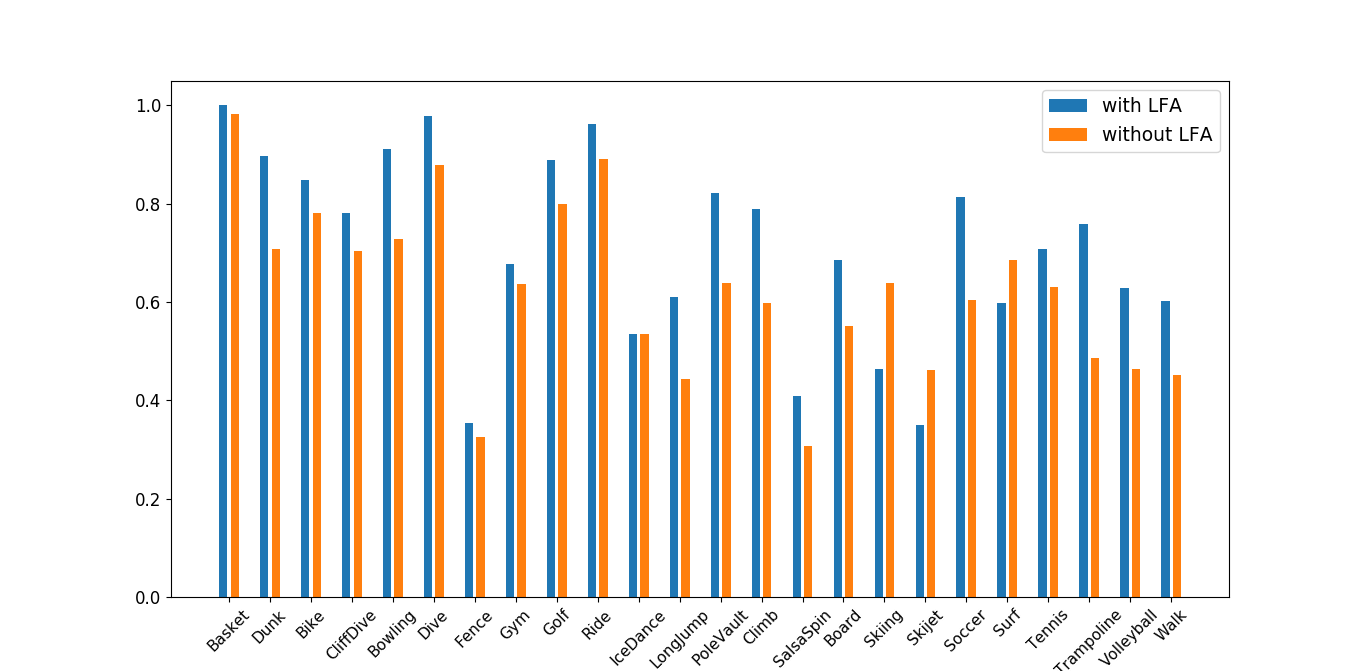}
\caption{Per-class AP on the UCF101-24 dataset with and without LFA module.}\label{fig:perclass}
\end{figure*}

\subsection{Ablation studies}

We conduct ablation studies on the UCF101-24 dataset. Note that all experiments are using only RGB data input.

\textbf{Long-term feature augmentation.} We first analyze the impact of the LFA since it is the key operation for long-term information integration. For comparison, we remove the module from our model and directly sample from the original 3D feature. In Table~\ref{tab:long}, the performance of our framework dropped $8.9\%$ in video-mAP without the long-term integration, which indicates that the temporal context is essential to final sample-wise detection. 

\lyx{In Fig.~\ref{fig:perclass}, we show the per-class AP value at the threshold of $\Delta=0.3$ on UCF101-24 dataset. To make comparison, we also reported results with the removal of LFA module. It can be observed for most action classes, LFA module can boost the detection performance. Especially, for some short-term ambiguous classes like ``Long Jump'' versus ``PoleVault'' and ``Cricket Bowling'', the enhancement is more obvious. (increased by $38\%, 29\%$ and $25\%  $ respectively.)}


\textbf{Embedding and shift vectors.} We also study how the embedding and shift vectors from association path helps action localization. In this experiment, we 
experimented with the removal of different vector components (and combinations of them)
while training with the remaining leftover parts.
For experiments without both components, we adopt a score-based strategy which always chooses the box with largest score in sample. The results are shown in Table ~\ref{tab:long}. We find that when the embedding vector is ignored, the result dropped $2.2\%$. Meanwhile, ignoring the potential shift vector resulted in drop of $1.5\%$ v-mAP. On the other hand, we find the baseline method with the largest score strategy performs \lyx{worse} than our method. These observation indicates that both the embedding and shift information are helpful towards the final box association process.

\begin{table}[htb]
    \centering
    \begin{tabular}{c|ccccc}\hline
         LFA &  & \checkmark & \checkmark & \checkmark & \checkmark \\
         Embedding & \checkmark & & & \checkmark & \checkmark \\
         Shift & \checkmark & & \checkmark &  & \checkmark \\ \hline
         v-mAP@0.3 & 62.2 & 67.4 & 68.9 & 69.6 & \textbf{71.1} \\ \hline
    \end{tabular}
    \caption{Results of applying the LFA module and association path vectors from the spatial detection network.}
    \label{tab:long}
\end{table}
\begin{table}[htb]
    \centering
    \begin{tabular}{c|c|c}\hline
         scheme & v-mAP@0.3 & per-video time (s) \\\hline
         fixed point-avg  &  67.4  & 0.551  \\\hline
         fixed point-4    &  63.3   &  0.412 \\\hline
         fixed point-10   &  67.5     &  0.654 \\\hline
         fixed step-avg   &   66.9    &  0.571 \\\hline
         DTS            & \textbf{71.1}  & \textbf{0.569} \\ \hline
    \end{tabular}
    \caption{Results of different sampling schemes}
    \label{tab:sample}
    \vspace{-3mm}
\end{table}

\begin{table*}[hbt]
    \small
    \centering
    \begin{tabular}{c|c|cc|cc|cc}\hline
         method & input  & \multicolumn{2}{c|}{JHMDB-21} & \multicolumn{2}{|c|}{UCFSports} &\multicolumn{2}{|c}{UCF101-24} \\\hline
        $\Delta$& & 0.2 & 0.5 & 0.2 & 0.5 & 0.3 & 0.5 \\\hline
        \cite{saha2016deep}&  RGB+Flow & 72.6  & 71.5& - & - & 54.9 & 35.9 \\\hline
        \cite{peng2016multi}&  RGB+Flow & 74.3 & 73.1 & 94.8 & 94.7 & 65.7 & 30.9 \\\hline
        \cite{kalogeiton2017action}&  RGB+Flow & 74.2 & 73.7 & 92.7 & 92.7  & - & 51.4 \\\hline
        \cite{hou2017end} & RGB+Flow & 78.4 & 76.9 & 95.2 & 95.2 & 69.4 & - \\ \hline
        \cite{singh2017online} &  RGB+Flow & 73.8 & 72.0 & - & - & - & 46.3 \\ \hline
        \cite{yang2017spatio}& RGB+Flow & - & - & - & - & 60.7 & 37.8 \\\hline
        \cite{song2019tacnet}& RGB+Flow & 74.1 & 73.4 & - & - & - & 52.9 \\\hline
        \cite{zhao2019dance}& RGB+Flow & - & 58.0 & - & 92.7 & - & 48.3 \\\hline
        \cite{li2018recurrent} & RGB+Flow & \textbf{82.3} & \textbf{80.5} & \textbf{97.8} &  \textbf{97.8} & \textbf{70.9} & - \\ \hline\hline
        \cite{saha2016deep} & RGB & 52.9 & 51.3 & - & - & 48.3 & 30.7 \\\hline
        \cite{saha2017amtnet}& RGB & 57.8 & 55.3 & - & - & 51.7 & 33.0 \\\hline
        \cite{li2018recurrent} & RGB & - & 61.7 & - & 87.6 & - & - \\\hline
        Ours & RGB & \textbf{76.1} & \textbf{74.3} & \textbf{94.3} & \textbf{93.8} & \textbf{71.1} & \textbf{54.0} \\\hline
       
    \end{tabular}
    \caption{Comparison with state-of-the-art methods (video-mAP), `-' denotes that the result is not available}
    \label{tab:sota}
    \vspace{-6mm}
\end{table*}

\textbf{Dynamic temporal sample}. To demonstrate the effectiveness of our dynamic temporal sample module, we conduct an experiment to compare between different sample strategies. (\romannumeral1) Fixed point sampling: For each temporal proposal, we sample a fixed number of 2D features for spatial detection. For comparison, we adopt the average sample number of algorithm \ref{alg:dynamic}; we denote this scheme with suffix \emph{`avg'}. We also test the performance with a larger ($10$) and smaller ($4$) number of samples, and they are denoted with suffixes \emph{`10'} and \emph{`4'} respectively. (\romannumeral2) Fixed step sampling: Given the normalized interval $(s, e)$, we sample features according to a fixed step size. Here, we set the step size as the average sample frequency of Algorithm \ref{alg:dynamic} during experiments, which is denoted with suffix \emph{`avg'}. (\romannumeral3) The proposed dynamic temporal sampling \emph{DTS} strategy.

Table~\ref{tab:sample} shows the experimental results with different sampling schemes. By comparison, we make the following observations: (\romannumeral1) By comparing \emph{DTS} and \emph{fixed point-avg}, we find that the two scheme achieved similar time costs, while \emph{DTS} outperformed the latter strategy in v-mAP terms. This is because \emph{fixed point-avg} is not adaptive to different dynamic levels, thus some samples could be redundant for simple action or scarce for complex actions. (\romannumeral2) \emph{fixed point-10} \lyx{performed worse} than \emph{DTS} and took more time to process the video, which shows that our strategy is more efficient \lyx{and effective}. 
On the other hand, \emph{fixed point-4} is worser than its 10-point counterpart, which indicates that less number of samples could be harmful to final tube detection. (\romannumeral3) \emph{DTS} and \emph{fixed step-avg} both consumes similar time costs, but \emph{DTS} outperformed \emph{fixed step-avg}. This indicates that being adaptive to the duration of action is less optimal than being adaptive to the complexity of the action. 



\subsection{Comparison with state-of-the-art}

In Table \ref{tab:sota}, we compare the detection performance of our framework with other state-of-the-art methods on three benchmark datasets. Among these datasets, our approach achieved competitive v-mAP results using only RGB image as input. In addition, we also compare our approach with some of recent methods that relied only on RGB input.

On UCF101-24, our method outperforms \cite{saha2017amtnet}, which is another RGB-only method, at all tested thresholds in this table. Compared with other methods that utilize both RGB and optical flow inputs, our methods \lyx{also achieves state-of-the-art results at both $\Delta=0.3$ and $\Delta=0.5$.} This indicates that our framework is effective and competitive even when the actions contain large motion and length variations. On the UCFSports and JHMDB-21 datasets, 
our method is also able to outperform \lyx{many recent} works~\cite{saha2017amtnet,kalogeiton2017action,singh2017online,zhao2019dance}. Overall, the comparisons on these three datasets indicate that our approach may be suited for \lyx{both short trimmed clips and long videos of variable length.} 
Although there is still a margin between our method and state-of-the-art \lyx{ConvNet + LSTM method} \cite{li2018recurrent} on UCF-Sports and JHMDB, their improvement mainly comes from the introduction of flow data, which is not involved in our work due to the efficiency consideration. This is most obvious
from the fact that our approach 
is superior 
to \cite{li2018recurrent} solely on RGB input. 


\subsection{Inference time}

\begin{figure}[]
    \centering
    \includegraphics[width=0.4\textwidth]{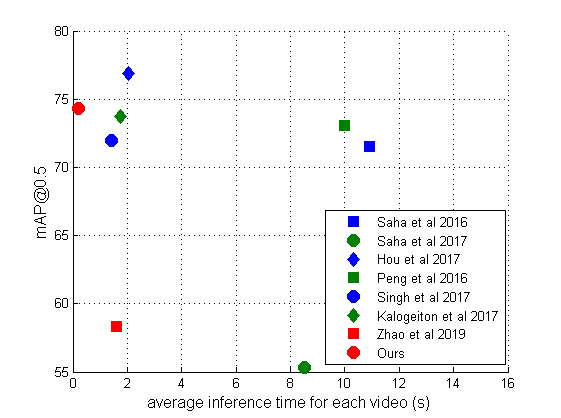}
    \caption{Time cost comparison between different pipelines}
    \label{fig:time}
    \vspace{-3mm}
\end{figure}

We also report the runtime cost for action tube detection. Following \cite{saha2016deep}, we run our model with data from JHMDB-21 for one epoch and compute the average inference time.
The result is reported in Fig.~\ref{fig:time}. Our framework infers each video in only $0.21$ seconds on average, achieving at most $7.6 \times$ faster than the nearest competitor. \lyx{In terms of FPS, our approach runs at approximately $168$ FPS, faster than other works reported in FPS speed \cite{singh2017online,kalogeiton2017action,yang2019step}}. 


\section{Conclusion}
In this paper, we depart from previous pipelines to propose a new framework for spatial-temporal action detection. We design the long-term augmentation mechanism and dynamic temporal sampling module to facilitate the detection process. We demonstrate the effectiveness of our approach on benchmarks of UCF101-24, JHMDB-21 and UCFSports. Besides, our model can process at a quicker speed, about $7.6\times$ faster than the nearest competitor.

\section{Acknowledgement}
The paper is supported in part by the following grants: China Major Project for New Generation of AI Grant(No. 2018AAA0100400), National Natural Science Foundation of China (No. 61971277, 61921066), CREST Malaysia Grant T03C1-17, and Adobe Research Gift.

{\small
\bibliographystyle{aaai.bst}
\bibliography{egbib.bib}
}

\end{document}